\documentclass[letterpaper]{article} % DO NOT CHANGE THIS
\usepackage{aaai24}  % DO NOT CHANGE THIS
\usepackage{times}  % DO NOT CHANGE THIS
\usepackage{helvet}  % DO NOT CHANGE THIS
\usepackage{courier}  % DO NOT CHANGE THIS
\usepackage[hyphens]{url}  % DO NOT CHANGE THIS
\usepackage{graphicx} % DO NOT CHANGE THIS
\usepackage{natbib}  % DO NOT CHANGE THIS AND DO NOT ADD ANY OPTIONS TO IT
\usepackage{caption} % DO NOT CHANGE THIS AND DO NOT ADD ANY OPTIONS TO IT
\usepackage{algorithm}
\usepackage{algorithmic}
\makeatletter

\DeclareRobustCommand\onedot{\futurelet\@let@token\@onedot}
\def\@onedot{\ifx\@let@token.\else.\null\fi\xspace}
\def\eg{\emph{e.g}\onedot} 
\def\sie{\textit{\textbf{SciEx}}\xspace}

\usepackage{xspace}

\def\eg{{e.g.,}\@\xspace}
\makeatother
\usepackage{newfloat}
\usepackage{svg}
\usepackage{amsmath}
\usepackage{makecell} 
\usepackage{booktabs}
\usepackage{subcaption}

\usepackage{listings}
\DeclareCaptionStyle{ruled}{labelfont=normalfont,labelsep=colon,strut=off} % DO NOT CHANGE THIS
\floatstyle{ruled}
\newfloat{listing}{tb}{lst}{}
\floatname{listing}{Listing}
\title{Exploring LLMs for Scientific Information Extraction\\ using the \sie Framework}
\author {
    Sha Li\equalcontrib \textsuperscript{\rm 1},
    Ayush Sadekar\equalcontrib \textsuperscript{\rm 1},
    Nathan Self\textsuperscript{\rm 1},
    Yiqi Su\textsuperscript{\rm 1},
    Lars Andersland\textsuperscript{\rm 2},\\
    Mira Chaplin\textsuperscript{\rm 2},
    Annabel Zhang\textsuperscript{\rm 1},
    Hyoju Yang\textsuperscript{\rm 2},
    James B Henderson\textsuperscript{\rm 2},\\
    Krista Wigginton\textsuperscript{\rm 2},
    Linsey Marr\textsuperscript{\rm 1},
    T.M. Murali\textsuperscript{\rm 1},
    Naren Ramakrishnan\textsuperscript{\rm 1}
}

\affiliations {
    \textsuperscript{\rm 1}Virginia Tech\\
    \textsuperscript{\rm 2}University of Michigan\\[1ex]
    \textsuperscript{\rm 1}\{shal, ayushs, nwself, yiqisu, ayzhang, lmarr, tmurali, naren\}@vt.edu \\
    \textsuperscript{\rm 2}\{landersl, mchaplin, hyojuya, jbhender, kwigg\}@umich.edu
}

\usepackage{bibentry}
\begin{document}

\maketitle

\begin{abstract}
Large language models (LLMs) are increasingly touted as powerful tools for automating scientific information extraction. However, existing methods and tools often struggle with the realities of scientific literature: long-context documents, multi-modal content, and reconciling varied and inconsistent fine-grained information across multiple publications into standardized formats. These challenges are further compounded when the desired data schema or extraction ontology changes rapidly, making it difficult to re-architect or fine-tune existing systems. We present \sie, 
a modular and composable framework that decouples key components including PDF parsing, multi-modal retrieval, extraction, and aggregation. This design streamlines on-demand data extraction while enabling extensibility and flexible integration of new models, prompting strategies, and reasoning mechanisms. We evaluate \sie on datasets spanning three scientific topics for its ability to extract fine-grained information accurately and consistently. 
Our findings provide practical insights into both the strengths and limitations of current LLM-based pipelines.
\end{abstract}

\section{Introduction}
\label{sec:intro}
Scientific information extraction
is the process of compiling
structured knowledge such as experimental parameters, relations, and outcomes from free-text
publications. An earlier generation of
handcrafted NLP tools for 
named entity recognition (NER), relation extraction (RE), and event extraction (EE) has given way to large language model (LLM)-driven
pipelines. However, the remarkable success of LLM
tools in general NLP tasks has not translated to improvements in
scientific information extraction.

There are many reasons for this disparity. 
First, scientific knowledge is distributed across heterogeneous modalities (text, tables, figures), requiring cross-modal reasoning to capture dependencies between methods, results, and interpretations. Second, concepts appear under diverse lexical and unit variations (\eg ``SARS-CoV-2 persistence” vs. ``COVID-19 virus viability”, molarity vs. ppm), which violate the assumptions of schema-constrained extractors. Third, scientific papers exhibit complex discourse structures, with evidence scattered across distant sections (\eg methods, results, supplementary materials). Traditional extractors, optimized for local or sentence-level context, cannot effectively aggregate such cross-document dependencies.

Current approaches to applying LLMs to
scientific information extraction use fine-tuning~\cite{dunn2022structured},  
prompt engineering~\cite{polak2024extracting, da2024automated}, or in-context learning. However, despite promising results, building stable and generalizable pipelines remains difficult due to prompt sensitivity, terminological and numerical inconsistency, and long-document dependencies. Scientific papers frequently exceed model context windows and distribute related evidence across multiple modalities, challenging current architectures to integrate this information coherently.

\iffalse
Despite showing promising results, applying LLMs to on-demand scientific information extraction remains challenging: \textbf{(1) Prompt sensitivity and instability:} current works depend heavily on prompt engineering and few-shot exemplars, which needs to be specially designed and curated for different LLMs, domains and tasks. This dependency also introduces variability and limits generalizability. \textbf{(2) Terminological and numerical inconsistency}: Technical terms, entity mentions, and numerical expressions often vary across paragraphs or publications, requiring domain-specific normalization (\eg, converting between molarity and ppm). Such inconsistencies hinder accurate entity linking and quantitative comparison. \textbf{(3) Long-document context and dependency}: Scientific papers frequently exceed LLM context windows and contain non-local references with dense, interdependent information spread across sections. Existing methods either struggle to capture these distributed dependencies or to integrate multi-modal evidence (from text, tables, and figures) into a coherent understanding.
\fi

To address these challenges, we implement \sie,
a framework that enables on-demand synthesis of structured knowledge from multiple scientific publications, transforming unstructured textual and visual content into structured formats. The framework converts unstructured PDFs into structured outputs, providing flexibility to incorporate and evaluate different LLMs, prompting strategies, and reasoning paradigms.
To evaluate the effectiveness of LLMs at this task, we manually annotate a benchmark dataset of 143 papers from the medical and environmental sciences, annotated by PhD students in these fields.

The main contributions of this work are summarized as follows:
\begin{itemize}
    \item We detail a prompt-driven, retrieval-augmented framework for scientific information extraction that unifies text, tables, and figures from research publications into structured representations. Our approach focuses on methodological advances for adaptable and modular information extraction rather than a fixed system implementation.
    \item We formalize a modular and composable architecture that decouples core components (PDF parsing, retrieval, extraction, and aggregation) allowing independent replacement, upgrading, or integration of new capabilities. This design principle emphasizes extensibility and supports rapid experimentation with different LLMs, prompting strategies, and retrieval configurations.
    \item We construct a new scientific information extraction dataset. Experiments on this dataset reveal that existing LLMs exhibit performance degradation when applied to new scientific domains, highlighting the importance of domain adaptation and generalization in LLM-based scientific reasoning.
    % \item Our framework demonstrates how modular design and data-driven evaluation can streamline the creation of structured scientific knowledge bases, providing a practical path for accelerating scientific discovery and LLM evaluation in previously underexplored research domains.
\end{itemize}

\section{Framework Design}
\label{sec:method}
%This section elaborates on the architecture and workflow of \sie. First, we we start with an overview of \sie (\S\ref{sec:framework_overview}). Then, we detail the technical pipeline of each module () 

\begin{figure*}[t]
    \vspace{-0.2cm}
    \centering
    \includegraphics[width=\textwidth]{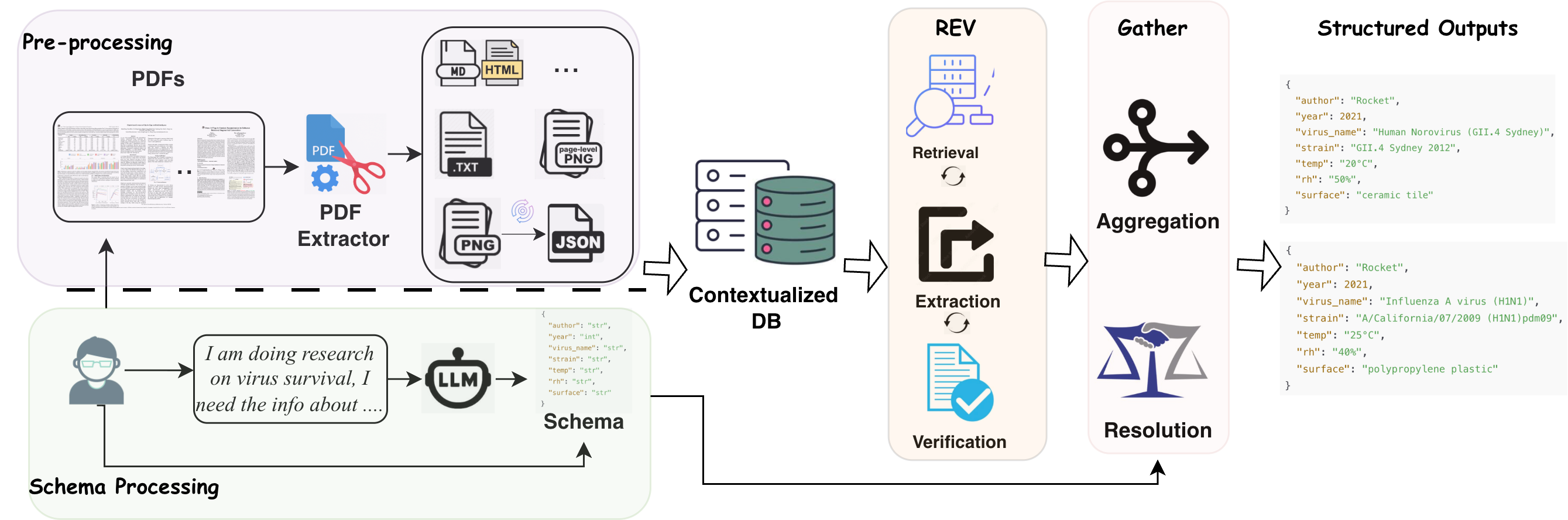}
    \caption{\textbf{Overall framework design and workflow of \sie.} \sie \textbf{pre-processes} PDFs using a \textit{PDF Extractor}, segmenting text, scientific figures and page-level images. Scientific figures are linked with captions, and all data points structured into JSON and stored in a \textit{contextualized database} with all extracted contents. Given a researcher’s request (explicit \textit{schema} or LLM-generated), the \textit{REV module} iteratively \textbf{retrieves, extracts, and verifies} information. An \textit{Aggregator} then \textbf{gathers} data from multiple PDFs, \textbf{canonicalizes} variant terms referring to the same entity, and \textbf{outputs} a unified, structured \textit{JSON}. }
    \label{fig:overview}
    \vspace{-0.5em}
\end{figure*}

%\subsection{Framework Overview}
\label{sec:framework_overview}
\sie is a prompt-driven and RAG-based framework for on-demand scientific information extraction, designed to transform unstructured research PDFs into structured, schema-conforming knowledge representations. The framework takes as input a collection of scientific publications together with user-specified information requirements, which can be provided either as explicit schemas or as natural language instructions describing the desired information.

Figure \ref{fig:overview} illustrates the overall architecture of \sie. The pipeline begins with the \textit{PDF Extractor}, which parses each PDF and segments its text, tables, and figures into standardized formats. The extracted content is stored in a \textit{Contextualized Database}, enabling efficient retrieval and downstream extraction. Researchers can define their information needs either by specifying structured schemas or by providing high-level natural language descriptions. Given a schema, the \textit{Retrieval-Extraction-Verification (REV) Module} performs recursive search and retrieval across the database, identifying relevant textual passages, tables, and figures. The module then applies LLM reasoning to extract information that aligns with the schema. Finally, the \textit{Aggregation Module} consolidates and standardizes extracted information from multiple publications, merging them into a unified, schema-conforming JSON representation.

Conceptually, the workflow can be interpreted as a distributed map-reduce operation: the PDF Extractor and REV modules perform a \emph{map} operation on each publication, applying the same extraction logic independently, while the Schema Aggregator acts as the \emph{reduce} operation, integrating outputs across documents to produce the final structured knowledge representation. 

The modular and extensible design of \sie allows flexible replacement, extension, and integration of off-the-shelf components. As a prompt-driven framework, \sie maintains adaptability by offering interfaces for researchers to incorporate domain-specific prompt designs when beneficial. In addition, \sie supports automatic prompt optimization via tools such as DSPy~\cite{khattab2023dspy}, which automate prompt engineering to enhance extraction performance.

\subsection{PDF Pre-processing}
\textbf{PDF Extractor}. The first step of \sie involves a \textit{PDF Extractor} parsing scientific publications in PDF format to extract both textual and visual content for downstream information extraction. In our implementation, we employ Docling\footnote{\url{https://docling-project.github.io/docling/}}, an open-source document conversion toolkit that performs fine-grained layout analysis and structural recognition~\cite{livathinos2025docling}.
 
For each paper, \sie extracts its textual content and segments it into multiple semantically coherent chunks. To retain visual information, all charts, diagrams, and tables are extracted as individual image files in PNG format while preserving their original spatial layout and visual fidelity. To ensure that only relevant scientific visual elements are included, a vision–language model (VLM) is employed for binary classification to distinguish scientific figures (\eg plots, charts, diagrams) from non-scientific illustrations (\eg logos or decorative images). Only validated scientific figures are preserved for subsequent data point extraction. Each scientific figure is paired with an associated caption, either extracted directly from the document when accurately detected or generated automatically by the VLM. The VLM also parses visual elements such as axis labels, legends, and data points (\eg bars, curves, markers), converting the figure into a structured JSON representation. This structured output is stored in the contextualized database, enabling joint reasoning over textual and visual modalities.

In addition, all pages of every paper are stored as full-page images, enabling \sie to jointly leverage textual, structural, and visual cues for comprehensive context understanding. The PDF Extractor also supports exporting extracted text and embedded images into structured formats such as HTML, XML or Markdown, facilitating flexible downstream integration and archival. 

The pre-processing pipeline supports both batch and parallel execution, ensuring scalability to large document collections and allowing for rapid iteration on pipeline improvements. Text segments are embedded and stored in a vector database for efficient semantic retrieval, while corresponding figures are indexed with metadata linking them to their source documents. Together, these components constitute a \textit{contextualized multi-modal knowledge base} that underpins the subsequent retrieval and information extraction stages.

\subsection{Schema Processing}
The \textit{Schema Module} defines the structured representation of the information to be extracted, serving as the interface between user intent and automated extraction. In \sie, a \textit{schema} specifies the desired attributes (\eg virus name, temperature, humidity, measurement units) and corresponding data types (\eg string, float, integer), which together organize the extracted knowledge into a consistent format. This schema-guided structure ensures uniformity and cross-paper consistency during information aggregation. Schemas can be provided in two modes: (1) explicit schema definition, where the user specifies key fields or entity–attribute pairs, or (2) implicit schema description, where the user provides a high-level instruction or query (\eg ``Extract virus survival durations under different environmental conditions"). For implicit cases, \sie then employs an LLM to generate a set of corresponding structured schema. In this way \sie supports domain experts who wish to impose precise data constraints and general users who rely on natural language prompts.

% This design allows \sie to balance flexibility and structure: it supports both domain experts who wish to impose precise data constraints and general users who rely on natural language prompts. By grounding the entire extraction process in schema-driven representation, \sie enables interpretable, reproducible, and semantically consistent scientific information extraction across diverse publication collections.

\subsection{Retrieval–Extraction–Verification}
Given the researcher-provided schema, the Retrieval–Extraction–Verification (REV) module iteratively discovers, extracts, and validates relevant information from the contextualized database. The process operates as a closed-loop pipeline that alternates between evidence retrieval, structured extraction, and verification until the extracted information reaches completeness and consistency criteria.

\textbf{Retrieval.} The schema defines the semantic intent and attribute types required for extraction. Using this schema as a query blueprint, the retriever searches the contextualized database to identify the top-$k$ most relevant evidence segments. These may include text chunks, table entries, or figure-derived JSON representations. The retrieval process leverages vector-based semantic search to match schema attributes with relevant content across modalities.

\textbf{Extraction.} Retrieved evidence is then passed to an LLM, which performs schema-guided extraction to produce structured records or tuples that conform to 
the schema. Schema-constrained decoding ensures that each extracted field adheres to the expected format and type (\eg numeric, categorical, or textual). The extracted results are serialized into an intermediate structured representation, such as JSON or a relational table, for downstream processing. Each extracted element is annotated with provenance metadata, including document identifiers, chunk indices, and figure references which ensures full traceability and verifiable linkage to the original context.

\textbf{Verification and Iteration.} To maintain factual accuracy and completeness, \sie performs self-verification over the extracted results. Missing, uncertain, or low-confidence fields trigger targeted follow-up queries that re-enter the retrieval stage, enabling iterative refinement. This closed-loop retrieval–extraction–verification cycle continues until convergence is achieved due to either the absence of missing fields, the attainment of a confidence threshold, or the completion of a predefined number of rounds. The iterative structure ensures that the final outputs are both semantically consistent and empirically grounded in the retrieved evidence.

\subsection{Aggregation and Resolution Module}
The \textit{Aggregation and Resolution} Module 
consolidates extracted information from multiple publications into a unified, schema-conforming representation. It aligns individual extraction results under the user-defined schema ensuring semantic consistency, unit normalization, and conflict resolution across heterogeneous data sources.

\textbf{Aggregation.} Records originating from different documents are grouped according to shared entities or experimental conditions (\eg identical virus strains, materials, or environmental parameters). Within each group, extracted fields are merged into composite entries. Numerical values reported in inconsistent units are automatically standardized using schema-defined normalization rules (\eg converting temperature values from Fahrenheit to Celsius), while categorical attributes are harmonized through controlled vocabularies. This aggregation step ensures that data from different studies can be seamlessly integrated and compared.

\textbf{Canonicalization and Conflict Resolution.} 
To enforce cross-publication coherence, \sie employs an LLM-based canonicalization process that maps lexical or morphological variants (\eg ``temp." and ``temperature") to consistent schema-defined terms. When conflicting values are encountered across sources, the system applies a hierarchical resolution strategy combining statistical and model-based verification. Specifically, cross-model ensembling and consistency voting are used to validate extractions: records are accepted only if corroborated by multiple prompts, model variants, or deterministic validation checks (\eg numeric range constraints, unit compatibility). Additionally, the LLM verifies several structural and factual integrity conditions:
(1) every extracted value must be traceable to its source context,
(2) list-type fields must contain complete and non-redundant value sets, and
(3) duplicate or contradictory entries within the same field are automatically removed. 

If a required piece of information is genuinely absent from all retrieved evidence, the corresponding field is explicitly labeled as \texttt{null}. This structured treatment of missing data maintains transparency and reliability in downstream analysis.

Overall, the Aggregation and Resolution module transforms fragmented, multi-source extractions into a coherent, verifiable, and structured knowledge representation, completing the information synthesis process within \sie.
 
\begin{figure}[H]
    \vspace{-0.2cm}
    \centering
    \includegraphics[width=0.5\textwidth]{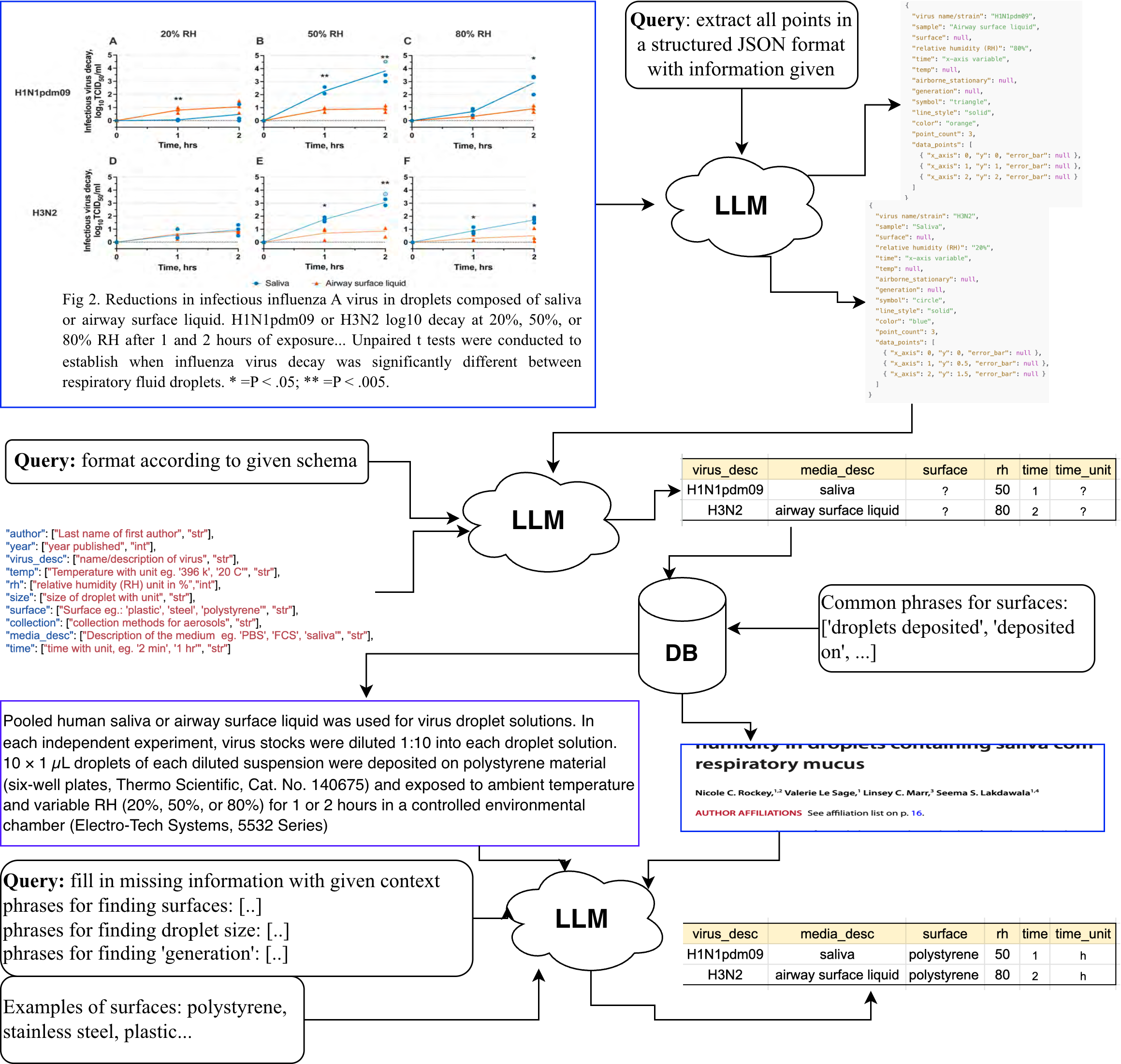}
    \caption{Processing of a single PDF through the \sie pipeline. Initial row data is extracted from relevant graphs in the PDF and formatted according to a specific schema. Missing information is identified and searched for through vector DB queries and passed examples to the LLM.}
    \label{fig:pipeline}
    \vspace{-0.5em}
\end{figure}

Figure~\ref{fig:pipeline} demonstrates how \sie works when applied to a sample paper. 

\section{Experiments}
\label{sec:exps}
\subsection{Dataset}
We evaluate \sie on three datasets spanning medical and environmental sciences.

\textbf{Virus Decay (VD)}. Publications in this dataset investigate how environmental factors and the surrounding medium influence the viability and infectivity of viruses.

\textbf{Ultraviolet (UV)} includes information used to calculate 224 UV disinfection rate constants for 59 viruses from 105 publications. All papers report UV disinfection of viruses in liquid suspension.

\textbf{Coagulation-Flocculation-Sedimentation (CFS)} includes 1,624 virus LRVs collected across 46 viruses from 43 eligible papers. All papers report CFS reduction of viruses in impure waters. The dataset includes 98 variables relating to water quality, process parameters, and virus reduction.

\subsection{Experiment Setup}
We primarily evaluate \sie using two large language models: Gemini-2.5-Flash~\cite{comanici2025gemini} and GPT-4o~\cite{hurst2024gpt}, applied for both textual and visual information extraction in our experiments. Importantly, the framework is model-agnostic; other LLMs, VLMs or MLLMs can be substituted for specific modality extraction tasks. For retrieval, we retrieve the top-5 most relevant chunks from the contextualized database each round.

\begin{figure}[H]
    \vspace{-0.2cm}
    \centering
    \includegraphics[width=\columnwidth]{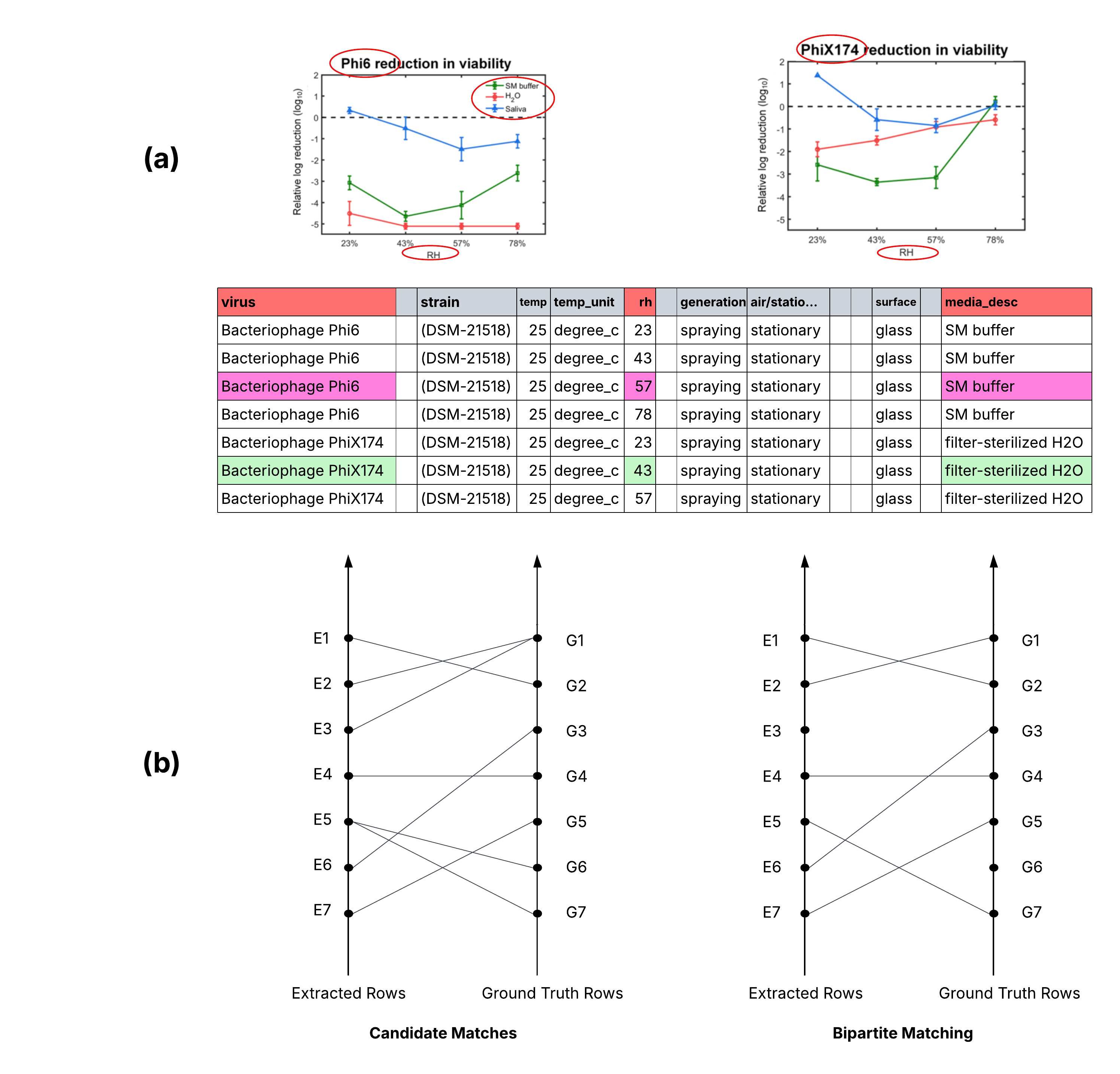}
    \caption{Row matching process. (a) Identify fields from ground truth that vary to distinguish each row. (b) Conduct a bipartite matching between candidate matches and ground truth rows.}
    \label{fig:matching_row}
    \vspace{-0.5em}
\end{figure}

\textbf{Evaluation Metrics.} We assess \sie’s performance using standard information extraction metrics, including precision, recall, F1-score, and accuracy across all datasets. 
These metrics collectively capture both the correctness and completeness of the extracted information.
To compare extracted outputs with ground truth, we perform row-level matching. Each ground-truth paper entry consists of multiple fields—typically representing experimental parameters or independent variables—which together define a unique record. For each ground-truth row, we identify candidate matches among the extracted rows based on field-level similarity. A bipartite matching algorithm is used to construct a mapping between the set of ground truth rows and the set of extracted rows. Figure~\ref{fig:matching_row} illustrates our row matching process for a single ground truth row.

\section{Results and Analysis}
\label{sec:results}
In this section, we present the experimental results and provide a detailed analysis of errors.

\subsection{Main Results}
We evaluate \sie's performance using Gemini-2.5-Flash and GPT-4o as the extraction models across three datasets. Performance is assessed along two dimensions: (1) completeness of extraction and (2) accuracy of extracted information. Table~\ref{tab:llm_comparison} summarizes the precision, recall, F1-score, and accuracy for each dataset.

\textbf{Cross-Dataset Analysis.} Both LLMs achieve higher performance on simpler datasets such as UV and VD, where most fields can be directly extracted from single figures or captions with minimal cross-referencing. In contrast, the more complex CFS dataset, which requires integrating information across multiple tables and figures, exhibits lower precision and F1-scores due to missing rows and challenges in reconciling distributed experimental conditions. Across datasets, recall generally exceeds precision, indicating that while extractions are often relevant, a substantial number of unwanted points are extracted. These results underscore the importance of iterative retrieval, verification, and multi-modal contextual reasoning to effectively handle heterogeneous and fine-grained scientific information.

\begin{figure}[H]
    \centering
    % First subfigure
    \begin{subfigure}[b]{0.22\textwidth}
        \centering
        \includegraphics[width=\textwidth]{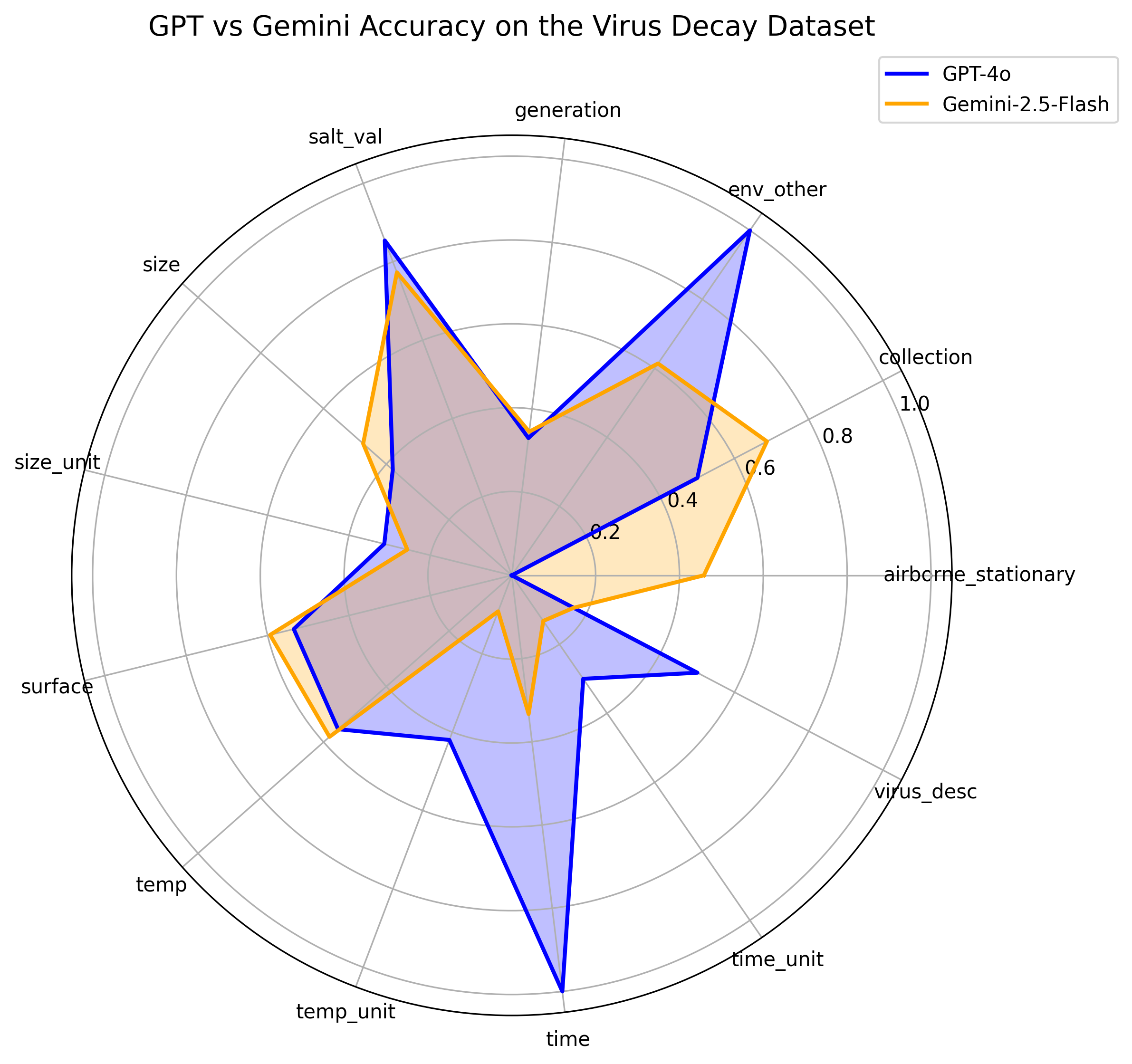}
        \caption{Virus Decay dataset}
        \label{fig:a}
    \end{subfigure}
    \hfill
    % Second subfigure
    \begin{subfigure}[b]{0.22\textwidth}
        \centering
        \includegraphics[width=\textwidth]{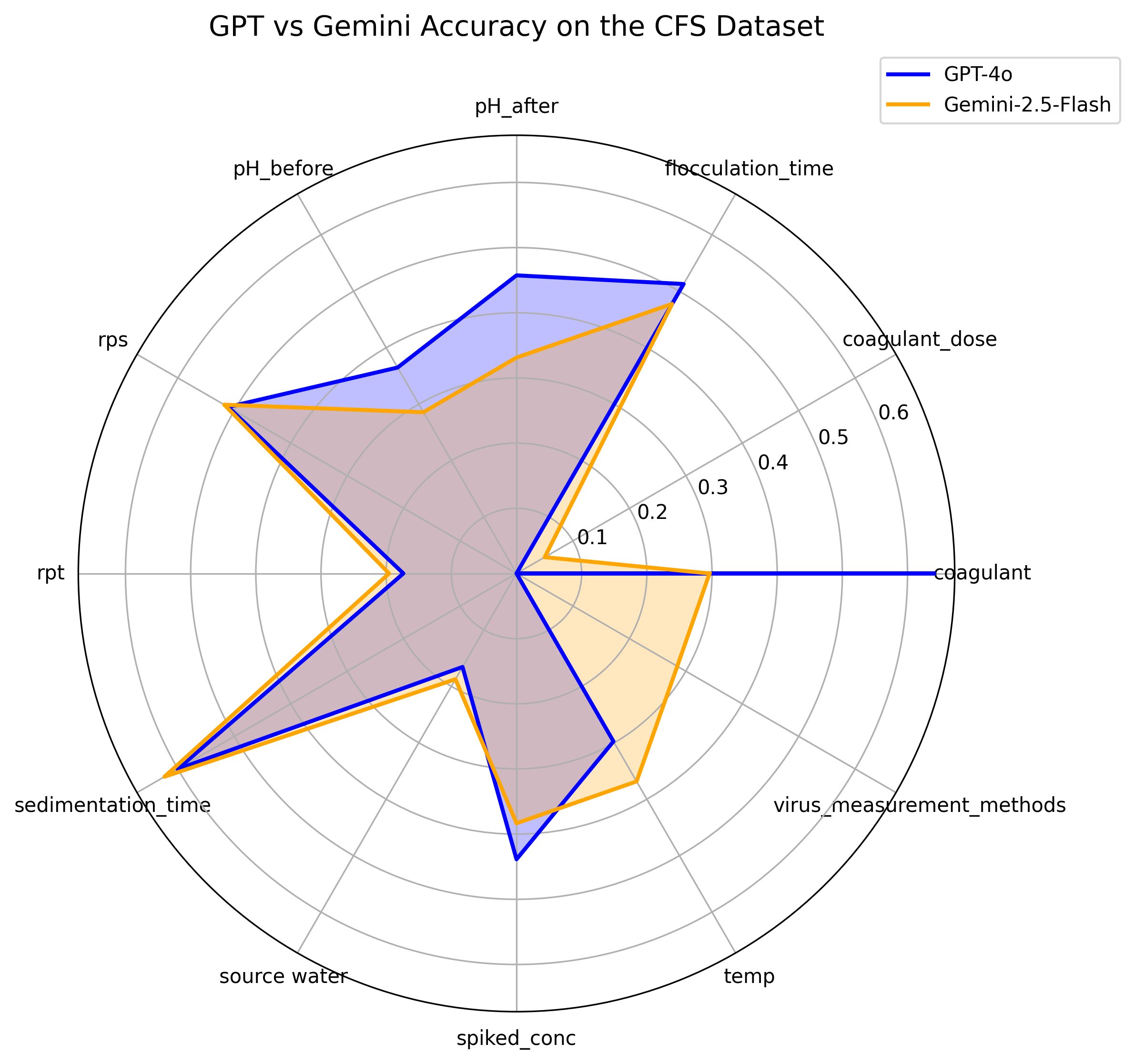}
        \caption{CFS dataset}
        \label{fig:b}
    \end{subfigure}
    % Reduce vertical space before main caption
    \vspace{-1mm}
    \caption{Accuracy of GPT-4o and Gemini-2.5-Flash on Virus Decay and CFS dataset. }
    \label{fig:accuracy}
\end{figure}

\textbf{Cross-Model Comparison.} GPT-4o consistently outperforms Gemini-2.5-Flash across all datasets, achieving higher average precision (0.26 vs. 0.22), recall (0.48 vs. 0.37), and F1-score (0.29 vs. 0.27). The gains are particularly notable for visually dominated datasets (UV and VD), reflecting GPT-4o's stronger multi-modal comprehension and identification capabilities. While Gemini-2.5-Flash is efficient, it tends to omit information, retrieving correct fragments but failing to fully populate schema fields, leading to lower recall.  Both models achieve moderate accuracy (0.5-0.6), suggesting that once a record is correctly localized, field-level extraction is generally reliable. Figure~\ref{fig:accuracy} compares accuracy of GPT-4o and Gemini-2.5-Flash on features of the Virus Decay (VD) and CFS datasets. Generally, GPT-4o demonstrates higher overall accuracy across most variables.

\begin{table}[t]
\centering
\caption{\sie's performance using (a) Gemini-2.5-flash and (b) GPT-4o across three datasets.}
\begin{subtable}[t]{0.45\textwidth}
\centering
\caption{Gemini-2.5-Flash}
\begin{tabular}{lcccc}
\toprule
\textbf{Dataset} & \textbf{Precision} & \textbf{Recall} & \textbf{F1-score} & \textbf{Accuracy} \\
\midrule
CFS   & 0.169 & 0.273 & 0.175 & 0.507 \\
UV    & 0.199 & 0.468 & 0.237 & 0.329 \\
VD & 0.284 & 0.382 & 0.297 & 0.556 \\
\bottomrule
\end{tabular}
\end{subtable}
\hfill
\begin{subtable}[t]{0.45\textwidth}
\centering
\vspace{-0.5mm}
\caption{GPT-4o}
\begin{tabular}{lcccc}
\toprule
\textbf{Dataset} & \textbf{Precision} & \textbf{Recall} & \textbf{F1-score} & \textbf{Accuracy} \\
\midrule
CFS   & 0.241 & 0.355 & 0.248 & 0.512 \\
UV    & 0.279 & 0.609 & 0.331 & 0.467 \\
VD & 0.333 & 0.476 & 0.380 & 0.580 \\
\bottomrule
\end{tabular}
\end{subtable}
\label{tab:llm_comparison}
\end{table}

\subsection{Error Analysis}
We manually examined representative examples from the three datasets to identify common error types and weaknesses in LLMs' ability to extract scientific information from PDFs. These analyses uncover key limitations of current LLM-based scientific IE systems and suggest promising directions for future improvement. We provide a detailed error analysis with examples in Appendix A. 

\textbf{Parsing and Document Quality.}  Older or poorly formatted PDFs pose substantial challenges due to degraded visual quality and inconsistent layouts. Graphs are often scanned or rasterized at low resolution, with truncated axes or illegible scales, which directly affects the LLM’s ability to extract precise numerical values. Improving preprocessing, such as resolution enhancement, adaptive binarization, or OCR refinement, may mitigate these issues.

\textbf{Text Extraction and Cross-Sentence Reasoning.} A frequent source of error arises from cross-sentence and cross-paragraph dependencies. When multiple entities (e.g., SARS-CoV-2 and Influenza A (H1N1)) appear in close proximity, the LLM can misattribute experimental conditions or outcomes, reducing relational accuracy. This emphasizes the need for discourse-aware and entity-grounded extraction mechanisms that maintain context while preventing misassignment across entities.

\textbf{Table Extraction and Structural Variability.} Extracting information from tables can be error-prone due to structural diversity and complexity. Variations in header hierarchies, units, and experimental parameters often cause schema mismatches and missing values. Related variables may be spread across multiple tables, which the current system cannot reconcile effectively. Structural irregularities reduce completeness of extracted fields. Addressing these challenges may require layout-aware parsing, hierarchical table modeling, and table-to-graph reasoning to link semantically related cells across tables.

\textbf{Figure Interpretation and Numeric Accuracy.} Extraction from figures introduces multiple sources of error.
Chart axes that employ implicit or inconsistent scales (linear- or log-based) produce numeric deviations. Dense visual clusters, such as overlapping curves or bar groups, lower recall as some points remain unrecognized.
Legends, titles, and captions are sometimes missing, either from loss of metadata during PDF-to-markdown conversion or because the source document assumes they can be inferred. This impairs contextual understanding, decreasing precision. Combining visual parsing (\eg via multi-modal pretraining for scale and symbol recognition) with contextual reasoning from surrounding text can help recover missing metadata and improve extraction accuracy.

\section{Related Work}
\label{sec:related_work}
Large language models (LLMs)~\cite{brown2020language, ouyang2022training} have advanced text understanding, reasoning, and generation, inspiring generative information extraction~\cite{zhang-etal-2025-survey}, where tasks like NER and RE are reformulated as sequence generation. Instruction-tuned and few-shot LLMs~\cite{wei2021finetuned, kojima2022large} enable flexible schema adaptation without retraining, while domain-specific variants~\cite{lee2020biobert, chithrananda2020chemberta, gu2021domain} enhance understanding of specialized corpora. RAG~\cite{lewis2020retrieval} further improves factuality through evidence grounding.

\textbf{LLMs for Scientific Literature.} Early work emphasized text-centric understanding through classification~\cite{vajjala2025text}, summarization~\cite{zhang2024comprehensive}, citation recommendation, and entity–relation extraction~\cite{huo2019semi}. Domain applications include materials science~\cite{olivetti2020data}, joint entity–relation extraction~\cite{dagdelen2024structured}, and medical IE using weak supervision~\cite{das2025weakly}. More recent studies demonstrate that LLMs achieve strong zero- and few-shot extraction without fine-tuning~\cite{ghosh-etal-2024-toward, da2024automated, woo2024evaluation, bhayana2024use}, while ChatExtract~\cite{polak2024extracting} enables conversational, iterative extraction.
Since much scientific knowledge resides in figures and tables, multi-modal understanding has gained attention. Prior work on figure analysis, chart captioning~\cite{hsu-etal-2021-scicap-generating, tang-etal-2023-vistext, yang2024scicap_plus, kantharaj-etal-2022-chart}, and table reasoning~\cite{zhou-etal-2025-m2} often relied on task-specific models with costly annotations. Recent multi-modal LLMs (MLLMs) such as mPLUG-DocOwl~\cite{hu2024mplug}, mPLUG-1.5~\cite{hu2024mplug1_5}, and UReader~\cite{ye-etal-2023-ureader} unify text–image reasoning in shared semantic spaces, though they remain limited to local contexts. OmniParser~\cite{wan2024omniparser} integrates text spotting, key information extraction, and table recognition, yet long-document reasoning and structured output generation remain challenging.

\textbf{LLM-powered Tools for Scientific Information Extraction}. 
A growing ecosystem of LLM-based tools supports interactive literature exploration. ChatPDF\footnote{\url{https://www.chatpdf.com/}} and ChatDoc\footnote{\url{https://www.chatdoc.com/}} enable conversational querying of papers. Elicit\footnote{\url{https://elicit.com/}} and SciSpace\footnote{\url{https://scispace.com/extract-data/}} automate summarization, data extraction, and multi-paper comparison. ScholarPhi~\cite{head2021augmenting} and Qlarify~\cite{fok2023qlarify} enhance interpretability via linked definitions and recursive exploration. LangExtract\footnote{\url{https://langextract.com/}} employs grounded extraction with visualization, while SciDaSynth~\cite{wang2025scidasynth} combines automated extraction with human validation for reliability.

These systems highlight the promise of LLMs for scalable and explainable scientific IE, but key challenges persist in cross-modal integration, schema generalization, and factual verifiability.
\section{Conclusion}
\label{sec:conclusion}
This paper presents \sie, an LLM-powered, prompt-driven, and RAG-based framework for on-demand scientific information extraction from a collection of scientific publications. Our framework addresses the limitations of existing methods by constructing a multi-modal contextualized database and employing an iterative retrieval–extraction–verification process to ensure the completeness and accuracy of fine-grained information. Our experiments on a dataset we built show that, while LLMs demonstrate promising capabilities, they are not yet ready for reliable large-scale deployment in scientific domains. Even with extensive prompt optimization, retrieval augmentation, and modular processing, the best results still fall short of the precision and completeness required for production-level knowledge extraction. These findings highlight several open challenges and opportunities for future research. In particular, further investigation is needed in (1) developing more robust domain adaptation and calibration strategies to improve generalization to unseen scientific areas, (2) enhancing cross-modal reasoning to better integrate textual, tabular, and visual information, and (3) establishing standardized datasets and evaluation protocols that more accurately reflect real-world scientific extraction tasks. 
\section{Acknowledgement}
\label{sec:ack}
This work is supported in part by US National Science Foundation grants DBI-2412389, CCF-1918770, CCF-1918770, and IIS-2312794. Any opinions, findings, and conclusions or recommendations expressed in this material are those of the author(s) and do not necessarily reflect the views of the sponsor(s).

\bigskip

%\newpage
\bibliography{aaai24}

\appendix

\section{Appendix A. Error Analysis}
\subsection{Error Type 1: Parsing and Document Quality}
\label{sec:error_old}
\begin{figure}[H]
    \vspace{-0.2cm}
    \centering
    \includegraphics[width=0.15\textwidth]{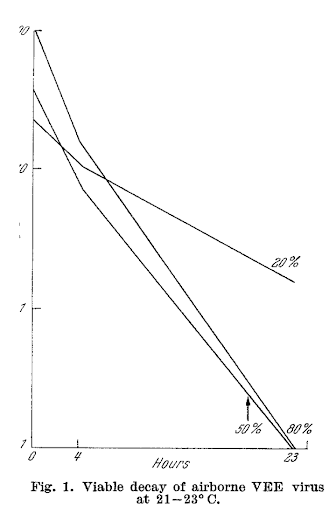}
    \caption{Older PDFs with ambiguous values}
    \label{fig:error_old}
    \vspace{-1em}
\end{figure}

Older or poorly formatted PDFs present significant challenges because of their degraded visual quality and inconsistent layouts. Graphs in these documents are frequently scanned or rasterized at low resolutions, often with truncated axes, unreadable scales or ambiguous values (\eg Figure \ref{fig:error_old}, image is from~\cite{harper1963influence}) which hinders the LLM's ability to accurately extract numerical data.

\subsection{Error Type 2: Inconsistent Table Structures}

\label{sec:error_table}
\begin{figure}[H]
        \vspace{-1em}
    \centering
    \includegraphics[width=0.4\textwidth]{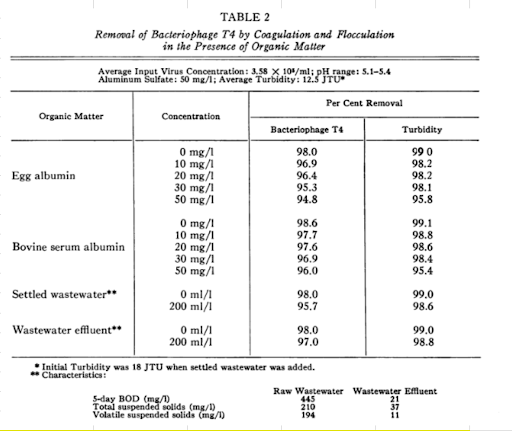}
    \caption{Example table from \cite{chaudhuri1970removal}}
    \label{fig:error_tb1}
    \vspace{-1em}
\end{figure}

\begin{figure}[H]
    \vspace{-0.2cm}
    \centering
    \includegraphics[width=0.4\textwidth]{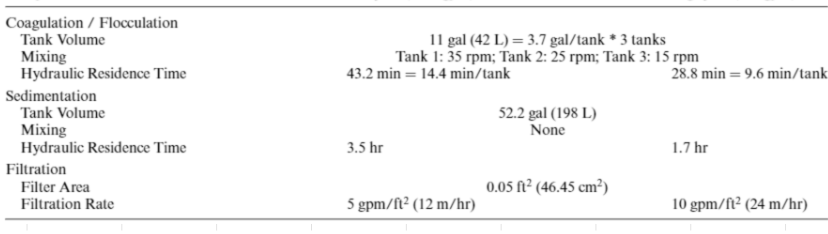}
    \caption{Example table from \cite{abbaszadegan2008removal}}
    \label{fig:error_tb2}
    \vspace{-1em}
\end{figure}

Tables (\eg Figure~\ref{fig:error_tb1} and Figure~\ref{fig:error_tb2}) often exhibit diverse structures, such as merged cells and nested headers. These variations make it challenging to identify relationships between rows and columns and to maintain a standardized representation of extracted data, leading to potential misalignment or loss of critical information during extraction.


\subsection{Error Type 3: Dense or Inconsistent Figures}
\label{sec:error_3}
\begin{figure}[H]
    \vspace{-0.2cm}
    \centering
    \includegraphics[width=0.4\textwidth]{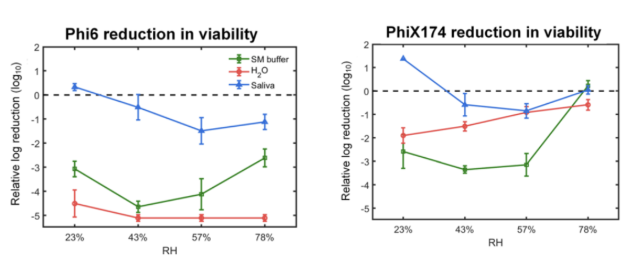}
    \caption{Single legend. Example figure from \cite{fedorenko2020survival}}
    \label{fig:error_3_legend}
    \vspace{-0.5em}
\end{figure}

\begin{figure}[H]
    \vspace{-0.2cm}
    \centering
    \includegraphics[width=0.4\textwidth]{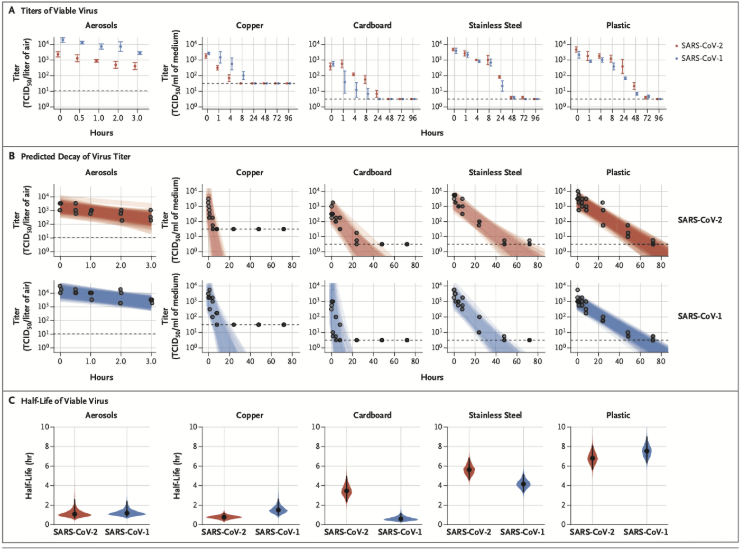}
    \caption{Large figure with many subfigures. Example figure from \cite{van2020aerosol}}
    \label{fig:error_3}
    \vspace{-0.5em}
\end{figure}

Figures containing multiple subplots or densely clustered visual elements pose substantial challenges for accurate information extraction. For example, in Figure~\ref{chang1985uv}, closely packed data points obscure individual values, causing LMs to overlook or merge points. In Figure~\ref{fig:error_3_legend}, grouped subfigures share a single legend that appears only once, making it difficult for LMs to correctly associate legend entries with their respective subfigures. In Figure~\ref{fig:error_3}, a large number of subfigures within a single image increases both visual and structural complexity, leading LMs to miss or partially extract information from certain subplots. Collectively, these issues impede precise parsing, correlation, and interpretation of visual information across complex figure layouts.
\begin{figure}[H]
    \vspace{-0.2cm}
    \centering
    \includegraphics[width=0.4\textwidth]{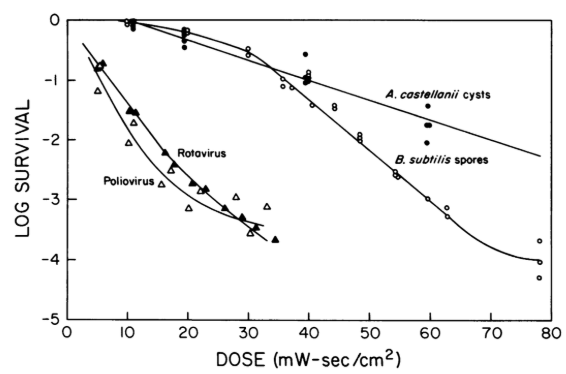}
    \caption{Clustered data points. Example figure from \cite{chang1985uv}}
    \label{fig:error_3_cluster}
    \vspace{-2em}
\end{figure}

\subsection{Error Type 4: Cross referencing}
\label{sec:error_4}
\begin{figure}[H]
    \vspace{-0.2cm}
    \centering
    \includegraphics[width=0.4\textwidth]{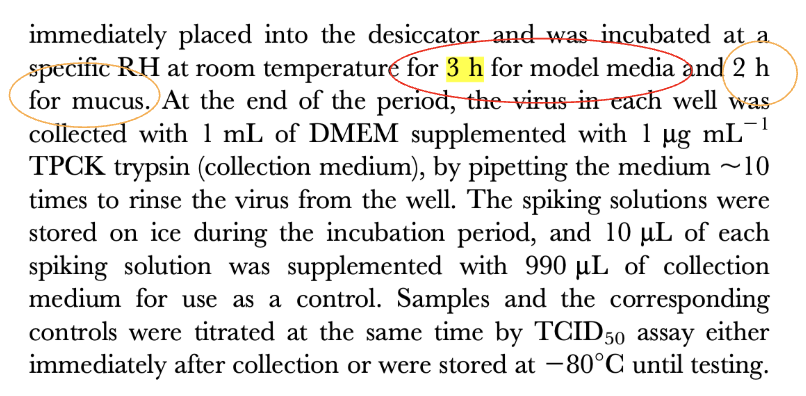}
    \caption{Cross referencing. Example from \cite{yang2012relationship}}
    \label{fig:error_4}
    \vspace{-0.5em}
\end{figure}

When figures, tables, or textual descriptions reference each other, LMs often struggle to establish correct correspondences between them. For example, in Figure~\ref{fig:error_4}, mucus-associated data could be mistakenly assigned to the 3h time point instead of 2h.

%\section{Appendix B. Illustration of pipeline workflow}

\end{document}